\documentclass[journal]{IEEEtran}
\usepackage{graphicx,url}
\usepackage{macro}
\usepackage{color}
\usepackage{pifont}
\newcommand{\cmark}{\ding{51}}%
\newcommand{\xmark}{\ding{55}}%
\begin{document}
\title{Deep Steering: Learning End-to-End Driving Model from Spatial and Temporal Visual Cues}

\author{Lu~Chi,
        and~Yadong~Mu,~\IEEEmembership{Member,~IEEE,}
\thanks{Lu Chi and Yadong Mu are with the Institute of Computer Science \& Technology, Peking University, China. E-mail: chilu@pku.edu.cn, myd@pku.edu.cn.}
}

\maketitle

\begin{abstract}
In recent years, autonomous driving algorithms using low-cost vehicle-mounted cameras have attracted increasing endeavors from both academia and industry. There are multiple fronts to these endeavors, including object detection on roads, 3-D reconstruction etc., but in this work we focus on a vision-based model that directly maps raw input images to steering angles using deep networks. This represents a nascent research topic in computer vision. The technical contributions of this work are three-fold. First, the model is learned and evaluated on real human driving videos that are time-synchronized with other vehicle sensors. This differs from many prior models trained from synthetic data in racing games. Second, state-of-the-art models, such as PilotNet, mostly predict the wheel angles independently on each video frame, which contradicts common understanding of driving as a stateful process. Instead, our proposed model strikes a combination of spatial and temporal cues, jointly investigating instantaneous monocular camera observations and vehicle's historical states. This is in practice accomplished by inserting carefully-designed recurrent units (e.g., LSTM and Conv-LSTM) at proper network layers. Third, to facilitate the interpretability of the learned model, we utilize a visual back-propagation scheme for discovering and visualizing image regions crucially influencing the final steering prediction. Our experimental study is based on about 6 hours of human driving data provided by Udacity. Comprehensive quantitative evaluations demonstrate the effectiveness and robustness of our model, even under scenarios like drastic lighting changes and abrupt turning. The comparison with other state-of-the-art models clearly reveals its superior performance in predicting the due wheel angle for a self-driving car.
\end{abstract}


\IEEEpeerreviewmaketitle

\section{Introduction}

\IEEEPARstart{T}he emerging autonomous driving techniques have been in the research phase in academia and in industrial R\&D departments for over decade. Level-3/4 autonomous vehicles are potentially becoming a reality in near future. Primary reasons for drastic technical achievement in recent years are a combination of several interlocking trends, including the renaissance of deep learning~\cite{Goodfellow-et-al-2016,JiaSDKLGGD14}, the rapid progression of devices used for sensing and in-vehicle computing, the accumulation of data with annotations, and technical breakthrough in related research fields (particularly computer vision). Over a large spectrum of challenging computer vision tasks (such as image classification~\cite{KrizhevskySH17} and object detection~\cite{DaiLHS16,RenHGZS17}), state-of-the-art computer vision algorithms have exhibited comparable accuracy to human performers under constrained conditions. Compared with other sensors like LIDAR or ultrasound, vehicle-mounted cameras are low-cost and can either independently provide actionable information or complement other sensors. For instance, one may expect these cameras to detect objects on the road (pedestrian, traffic signs, traffic light, obstacles in the front road) or estimate the orientation / distance of other cars, or even reconstruct 3-D dense maps of the surrounding environment.

Vision-based driver assist features have been widely supplied in modern vehicles. Typical features include collision avoidance by estimating front car distance, pedestrian / bicycle detection, lane departure warning, intelligent headlamp control etc. This research targets \emph{autonomous steering}, which is a relatively unexplored task in the fields of computer vision, robotics and machine learning. The goal is learning a vision-oriented model for autonomously steering a car. Unlike most prior deep models that primarily output static intermediate representations, the models developed in this work directly produce \emph{actionable} steering commands (accurate wheel angles, braking or acceleration etc.).

\begin{figure*}
\centering
\includegraphics[width=0.75\linewidth]{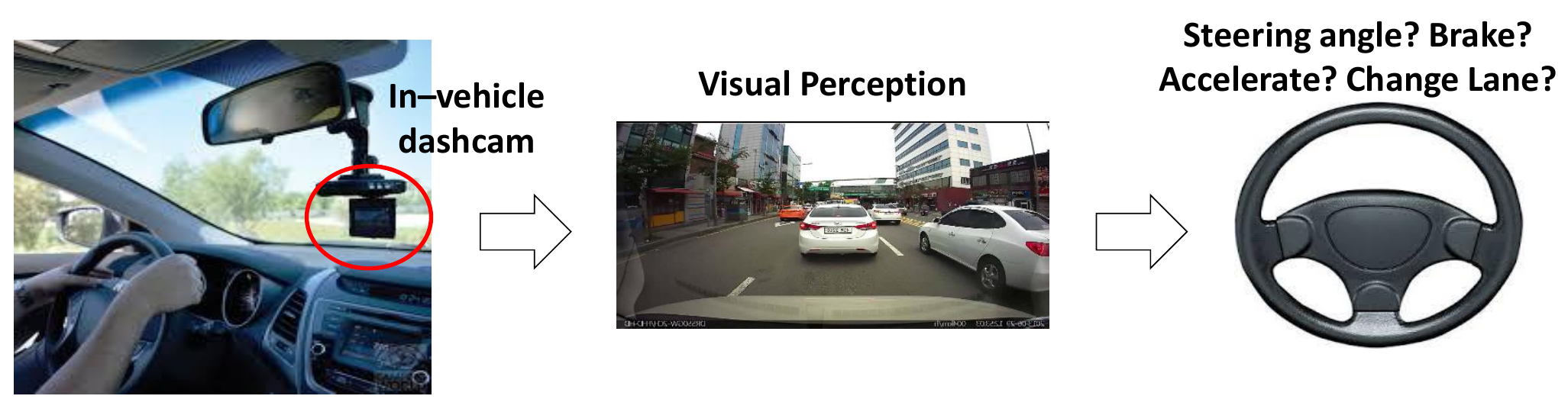}
\caption{Illustration of the application scenario of an end-to-end model that predicts instantaneous wheel angle or other steering operations.}
\label{fig:problem}
\end{figure*}

Generally, latest demonstration systems of autonomous steering adopt either a mediated perception approach~\cite{ChenSKX15} or behavior reflex approach~\cite{BojarskiTDFFGJM16,XuGYD16}, both of which have notably profitted from recent advances in deep learning. We postpone more detailed survey of these two paradigms in the Related Work section. This paper follows the paradigm of behavior reflex. Our proposed method, which we term Deep Steering, is motivated by the shortcomings in existing methods. The technical contributions offered by Deep Steering can be summarized as below:

First, most existing works train deep networks from images of a front-facing dashcam paired with the time-synchronized steering angle, which can be recorded from a human driver, electronic racing games~\cite{ChenSKX15} or estimated by IMU sensors~\cite{XuGYD16}. We argue that training from real-life driving logs is crucial to ensure the vehicle's safety when deploying the trained model in real cars. The data collected from racing game TORCS, for example, have biased distribution in visual background and road traffic and thus severely diverge from real driving scenarios. The work by Xu et al.~\cite{XuGYD16} builds a deep model using a subset of recently-established BDD-Nexar Collective\footnote{\url{https://github.com/gy20073/BDD_Driving_Model/}}, which contains about 300-hour video data. However, BDD-Nexar Collective records limited information besides video, mainly GPS and IMU. The driver's actions can be only indirectly estimated from the IMU information. The accuracy of wheel angle annotations and the synchronization of visual/non-visual information are not fully validated. Moreover, the major goal of Xu et al. is predicting discrete vehicle state (such as go straight or turn left) rather than continuous steering actions. In contrast, our work performs all training and model evaluation based on real high-quality human driving logs.

Second, existing methods mostly learn a model of steering actions from individual video frame. Intuitively, previous vehicle states and temporal consistency of steering actions play a key role in autonomous steering task. However, they are either completely ignored in the model-learning process~\cite{ChenSKX15,BojarskiTDFFGJM16} or inadequately utilized~\cite{XuGYD16}. In this work we explored different architectures of recurrent neural network. We empirically find that it is a better choice to simultaneously utilize temporal information at multiple network layers rather than any single layer. In practice, the idea is implemented by a combination of standard vector-based Long Short-Term Memory (LSTM) and convolutional LSTM at different layers of the proposed deep network..

Last but not least, deep models are conventionally regarded as complicated, highly non-linear ``black box". The prediction of these models, despite often highly accurate, is not understandable by human. In autonomous driving, safety is of highest priority. It is crucial to ensure the end users fully understand the mechanism of the underlying predictive models. There are a large body of research works on visualizing deep networks, such as the work conducted by Zeiler et al.~\cite{ZeilerF14} and global average pooling (GAP)~\cite{ZhouKLOT16}. This work adapts the visual back-propagation framework~\cite{BojarskiCCFJMZ16b} for visually analyzing our model. Salient image regions that mostly influence the final prediction are efficiently computed and visualized in an human-readable way.

The reminder of this paper is organized as follows. Section~\ref{sec:relatedwork} reviews the relevant works developed in the past years. Problem setting is stated in Section~\ref{sec:problem}. Our proposed Deep Steering is presented in Section~\ref{sec:model}. Comprehensive empirical evaluations and comparisons are shown in Section~\ref{sec:experiment} and this work is concluded in Section~\ref{sec:conclusion}.

\begin{figure*}[t]
\centering
   \includegraphics[width=0.9\linewidth]{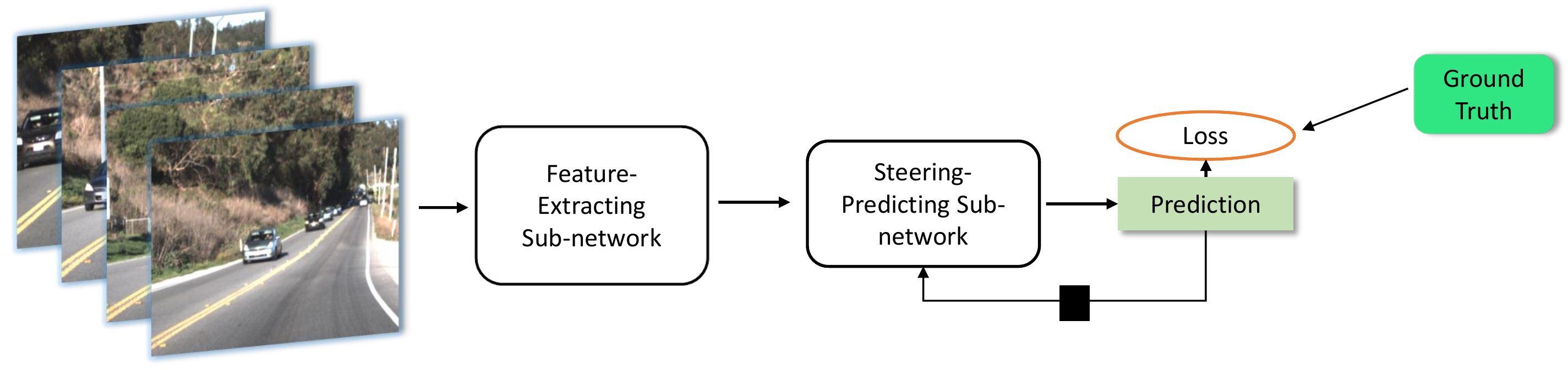}
   \caption{The architecture of our proposed deep networks for the task of vision-oriented vehicle steering. The arrows in the network denote the direction of data forwarding, and the black block on the arrow means a 1-step information delay in recurrent units. See main text for more explanation of these two sub-networks. Crucially, the steering-predicting sub-network reads predictions in previous step, including speed, torque and wheel angle.}
\label{fig:net}
\end{figure*}

\section{Related Work}
\label{sec:relatedwork}

The ambition of autonomous driving can trace back to Leonardo da Vinci's self-propelled cart if not the earliest, whose complicated control mechanism allows it to follow a pre-programmed path automatically. To date, self-driving cars are no longer a rare sight on real roads. The research of autonomous driving has received tremendous governmental funding such as Eureka Prometheus Project\footnote{\url{http://www.eurekanetwork.org/project/id/45}} and V-Charge Project\footnote{\url{http://www.v-charge.eu/}}, and was stimulated by competitions like DARPA Grand Challenge\footnote{\url{https://en.wikipedia.org/wiki/DARPA_Grand_Challenge}}.

Following the taxonomy used in~\cite{ChenSKX15}, we categorize existing autonomous driving systems into two major thrusts: \emph{mediated perception approaches} and \emph{behavior reflex approaches}. For the former category, this difficult task is first decomposed into several atomic, more tractable sub-tasks of recognizing driving-relevant objects, such as road lanes, traffic signs and lights, pedestrians, etc. After solving each sub-task, the results are compiled to obtain a comprehensive understanding of the car's immediate surroundings, and a safe and effective steering action can then be predicted. Indeed, most industrial autonomous driving systems can be labeled as mediated perception approaches. Vast literature on each afore-mentioned sub-tasks exists. Thanks to deep learning, we have witnessed significant advances for most sub-tasks. Particularly, object detection techniques~\cite{DaiLHS16,GirshickDDM14}, which locate interested object with bounding boxes (and possibly 3-D orientation), are regarded as key enablers for autonomous driving, including vehicle / pedestrian detection~\cite{DalalT05,ZhuYCA06,TuzelPM07,YangR13} and lane detection~\cite{Aly14,HuvalWTKSPARMCM15,GurghianKBCM16}.

The Deep Driving work by Xiao et al.~\cite{ChenSKX15} represents novel research in this mediated perception paradigm. Instead of object detection on roads, the authors proposed to extract affordance more tightly related to driving. Examples of such affordance include the distance to nearby lane markings, distance to the preceding cars in the current / left / right lanes, and angle between the car¡¯s heading and the tangent of the road. This way aims no waste of task-irrelevant computations. The authors devise an end-to-end deep network to reliably estimate these affordances with boosted robustness.

Regarding behavior reflex approaches, Dean Pomerleau developed the seminal work of ALVINN~\cite{Pomerleau88}. The networks adopted therein are ``shallow" and tiny (mostly fully-connected layers) compared with the modern networks with hundreds of layers. The experimental scenarios are mostly simple roads with few obstacles. ALVINN pioneered the effort of directly mapping image pixels to steering angles using a neural network. The work in~\cite{KoutnikCSG13} trained a ``large", recurrent neural networks (with over 1 million weights) using a reinforcement learning method. Similar to Deep Driving~\cite{ChenSKX15}, it also utilized TORCS racing simulator for data collection and model testing. The DARPA-seeding project known as DAVE (DARPA Autonomous Vehicle)~\cite{LeCunMBCF05} aims to build small off-road robot that can drive on unknown open terrain while avoiding obstacles (rocks, trees, ponds etc) solely from visual input. DAVE system was trained from hours of data by a human driver during training runs under a wide variety of scenarios. The network in~\cite{LeCunMBCF05} is a 6-layer convolutional network taking a left/right pair of low-resolution images as the input. It was reported that DAVE's mean distance between crashes was about 20 meters in complex environments. DAVE-2~\cite{BojarskiTDFFGJM16,BojarskiCCFJMZ16b} or PilotNet~\cite{BojarskiYCCFJM17} were inspired by ALVINN and DAVE. The network consists of 9 layers, including a normalization layer, 5 convolutional layers and 3 fully connected layers. The trained data is collected from two-lane roads (with and without lane markings), residential roads with parked cars, tunnels, and unpaved roads.

We would argue that temporal information has not been well utilized in all afore-mentioned work. For instance, PilotNet learned to control the cars by solely looking into current video frame. The Deep Driving work adopts another different approach. The authors uses some physical rules to calculate the due speed and wheel angles to obtain a smooth driving experience. However, in the case of curved lanes, Deep Driving tends to predict inaccurate wheel angle from the physical rules. The most relevant to ours is the work in~\cite{XuGYD16}. The authors insert an LSTM unit in the penultimate layer. LSTM's internal state is designed to summarize all previous states. The authors set 64 hidden neurons in LSTM, which we argue is inadequate to effectively capture the temporal dependence in autonomous driving. Our work enhances temporal modeling by exploring a mix of several tactics, including residual accumulation, standard / convolutional LSTM~\cite{ShiCWYWW15} at multiple stages of the network forwarding procedure. We validate on real data that the proposed network better captures spatial-temporal information and predicts more accurate steering wheel angle.

\section{Problem Formulation}
\label{sec:problem}

This section formally specifies the problem that we consider. We use hours of human driving record for training and testing a model. The major input is a stream of video frames captured by the front-facing camera installed in a car. In addition, during the training time, we are also provided with the instantaneous GPS, speed, torque and wheel angle. All above information is monitored and transmitted through the CAN (Controller Area Network) bus in a vehicle. More importantly, information from different modalities is accurately synchronized according to time stamps.

The central issue of this task is to measure the quality of a learned model for autonomous steering. Following the treatment in prior studies~\cite{LeCunMBCF05,BojarskiYCCFJM17}, we regard the behavior of human drivers as a reference for ``good" driving skill. In other words, the learned model for autonomous steering is favored to mimic a demonstrated human driver. The recorded wheel angles from human drivers are treated as ground truth. Multiple quantitative evaluation metrics that calculate the divergence between model-predicted wheel angles and the ground truth exist. For instance, the work of~\cite{XuGYD16} discretized the real-valued wheel angles into a fixed number of bins and adopt multi-class classification loss. In PilotNet, the training loss / evaluation criterion are different: model training is based on per-frame angle comparison, and the testing performance is evaluated by the counts of human interventions to avoid road emergence. Specifically, each human intervention triggers a 6-second penalty, and the ultimate testing performance is evaluated by the percentage of driving time that is not affected by human intervention. The major problem with PilotNet lies in that the testing criterion does not distinguish different levels of bad predictions (e.g., a deviation from the ground truth by $1^\circ$ or $20^\circ$ does make a difference).

\begin{figure*}
\centering
   \includegraphics[width=0.95\linewidth]{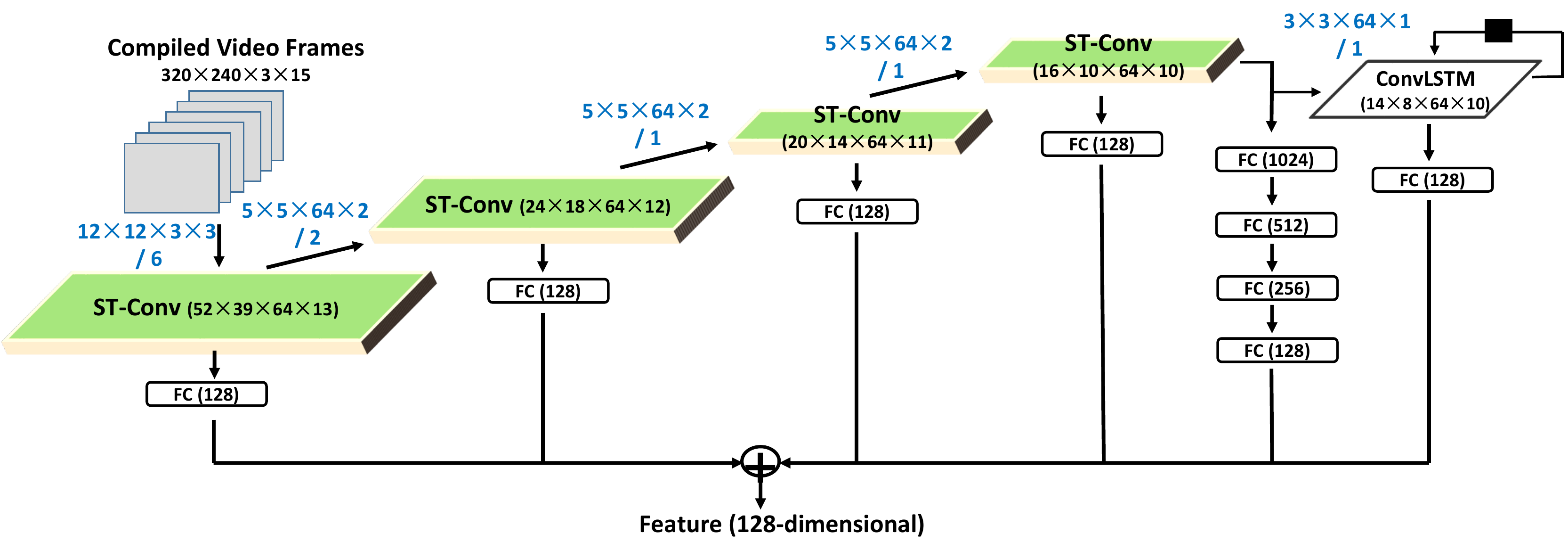}
   \caption{The design of feature-extracting sub-network. The proposed sub-network enjoys several unique traits, including spatio-temporal convolution (ST-Conv), multi-scale residual aggregation, convolutional LSTM etc. ReLu and DropOut layers are inserted after each ST-Conv layer for non-linear activation and enhancing generalization ability respectively. They are not displayed due to space limit. See main text for more details.}
\label{fig:conv}
\end{figure*}

We adopt a simple form of squared loss that is amenable to gradient back-propagation. The objective below is minimized:
\begin{eqnarray}
\mathcal{L}_{steer} = \frac{1}{T} \sum_{t=1}^T \left\| \tilde s_{t,steer} - s_{t,steer} \right\|^2, \label{eqn:steer}
\end{eqnarray}
where $s_{t,steer}$ denotes the wheel angle by human driver at time $t$ and $\tilde s_{t,steer}$ is the learned model's prediction.

\section{Network Design and Parameter Optimization}
\label{sec:model}

For statement clarity, let us conceptually segment the proposed network into multiple sub-networks with complementary functionalities. As shown in Fig.~\ref{fig:net}, the input video frames are first fed into a \emph{feature-extracting sub-network}, generating a fixed-length feature representation that succinctly models the visual surroundings and internal status of a vehicle. The extracted features are further forwarded to a \emph{steering-predicting sub-network}. Indeed, the afore-mentioned network decomposition is essentially conceptual - it is not very likely to categorize each layer as exclusively contributing to feature production or steering control.

\subsection{Feature-Extracting Sub-network}
\label{subsec:fen}

Fig.~\ref{fig:conv} provides an anatomy of the sub-network which majorly plays the role of extracting features. The sub-network is specially devised for our interested task. Here we would highlight its defining features in comparison with off-the-shelf deep models such as AlexNet~\cite{KrizhevskySH17}, VGG-Net~\cite{SimonyanZ14a} or ResNet~\cite{HeZRS16}.

\begin{figure}[t]
\centering
   \includegraphics[width=0.95\linewidth]{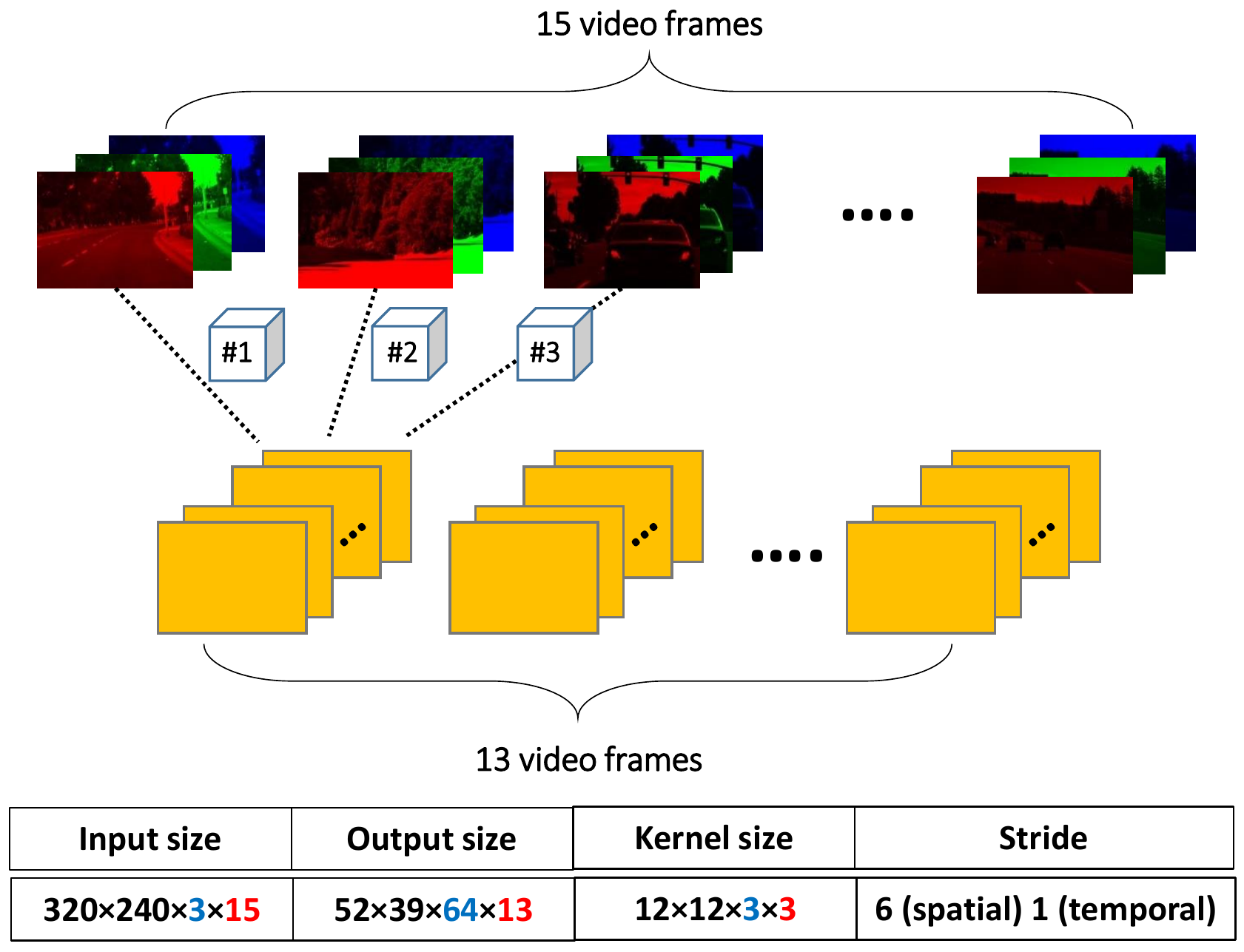}
   \caption{Illustration of Spatio-Temporal Convolution (ST-Conv) by convoluting the input 15-frame video clip with the first ST-Conv layer. Sizes related to channels / temporal information are highlighted in blue and red respectively.}
\label{fig:3dconv}
\end{figure}

\subsubsection{Spatio-Temporal Convolution (ST-Conv)}

Consecutive frames usually have similar visual appearance, but subtle per-pixel motions can be observed when optical flow is computed. Conventional image convolutions, as those adopted by state-of-the-art image classification models, can shift along both spatial dimensions in an image, which implies that they are essentially 2-D. Since these convolutions operate on static images or multi-channel response maps, they are incapable of capturing temporal dynamics in videos. Prior research on video classification has explored compiling many video frames into a volume (multiple frames as multiple channels) and then applying 2-D convolution on the entire volume. Though being able to encode temporal information, such a treatment has severe drawbacks in practice. The learned kernels have fixed dimensions and are unable to tackle video clips of other sizes. In addition, the kernels are significantly larger than ordinary 2-D convolution's due to the expanded channels. This entails more learnable parameters and larger over-fitting risk.

Inspired by the work of C3D~\cite{TranBFTP15}, we here adopt \emph{spatio-temporal convolution} (ST-Conv) that shifts in both spatial and temporal dimensions. Fig.~\ref{fig:3dconv} illustrates how the first ST-Conv layer works. As seen, the input are a compilation of 15 consecutive video frames. Each has RGB channels. Because the road lanes are known to be key cues for vehicle steering, we utilize a relatively large spatial receptive field ($16 \times 16$) in a kernel's spatial dimensions. And similar to AlexNet, a large stride of 6 is used to quickly reduce the spatial resolution after the first layer. To encode temporal information, convolutions are performed cross adjacent $k$ frames, where $k$ represents the parameter of temporal receptive field and here set to 3. We do not use any temporal padding. Therefore, apply a temporal convolution with width 3 will eventually shrink a 15-frame volume to 13.

\subsubsection{Multi-Scale Residual Aggregation}

In relevant computer vision practice, it is widely validated that response maps at many convolutional layers are informative and complementary to each other. Deep networks are observed to first detect low-level elements (edges, blobs etc.) at first convolutional layers, and gradually extend to mid-level object parts (car wheels, human eyes etc.), and eventually whole objects. To capitalize on cross-scale information, multi-scale aggregating schemes such as FCN~\cite{ShelhamerLD17}) or U-net~\cite{IsolaZZE16} have been widely adopted. In this work, we adopt ResNet-style skip connections, as shows in Fig.~\ref{fig:conv}. Specifically, we set the target dimension of the feature-extracting sub-network to be 128. Responses at ST-Conv layers are each fed into an FC (fully-connected) layer, which converts anything it received to a 128-d vector. The final feature is obtained by adding these 128-d vectors at all scales. Since the above scheme utilizes skip connections from the final feature to all intermediate ST-Conv layers, it is capable of mitigating the gradient vanishing issue, similar to ResNet.

\subsubsection{Convolutional LSTM}

\begin{figure}
\centering
   \includegraphics[width=\linewidth]{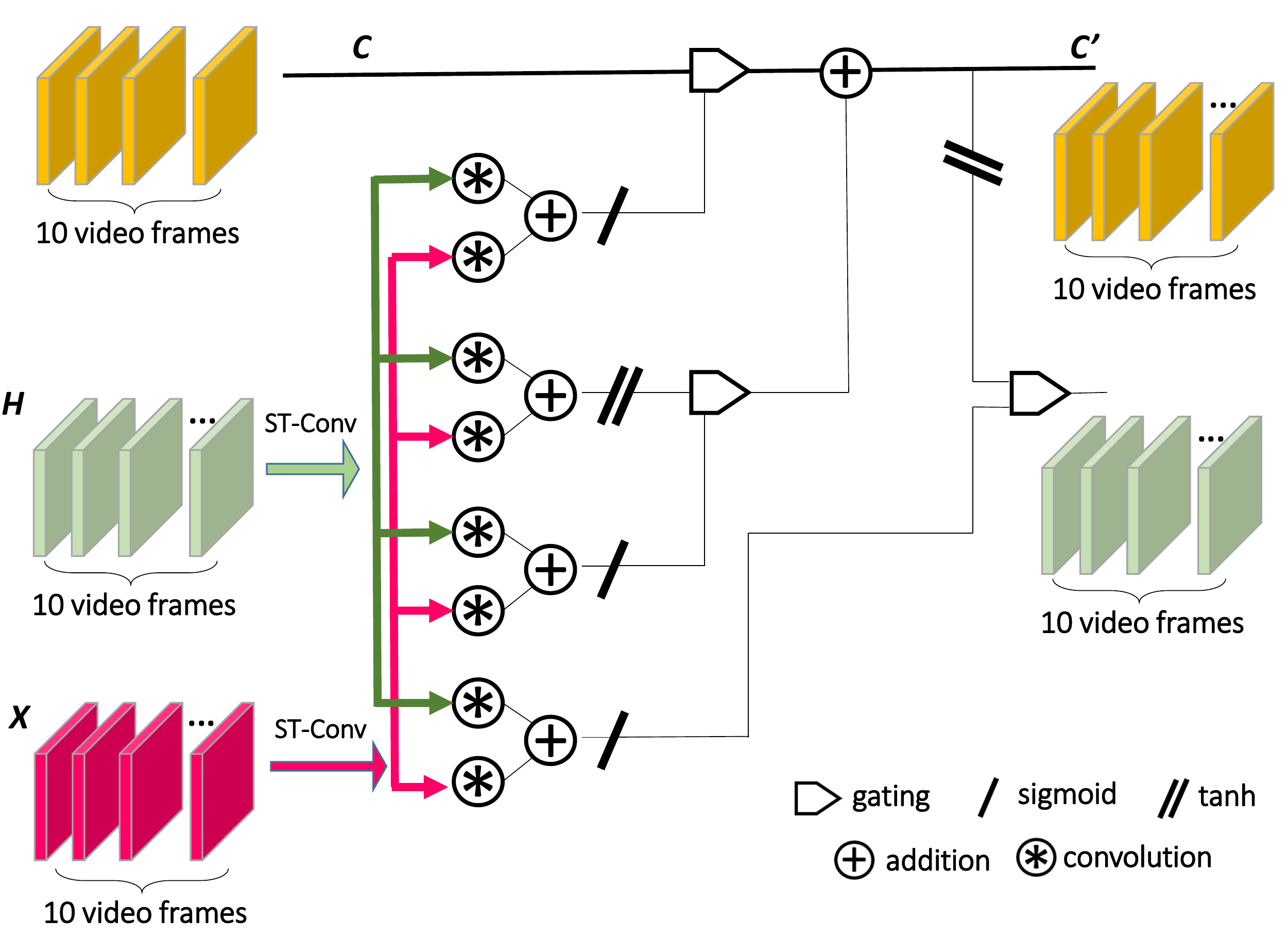}
   \caption{Data flow in Conv-LSTM. We replace the vector-to-vector multiplication in standard LSTM with spatio-temporal convolutions as described before.}
\label{fig:convlstm}
\end{figure}

Autonomous steering is intrinsically a sequential learning problem. For each wheel angle, it is determined both by the current state and previous states that the model memorizes. Recurrent neural network (such as LSTM) is one of the major workhorses for tackling such scenarios. LSTM layer is often inserted right before the final loss. We also introduce recurrent layers in the early feature-extracting stage.

For the problem we consider, the input and output of all ST-Conv layers are 4-D tensors: the first two dimensions chart the spatial positions in an image, the third indexes different feature channels and the fourth corresponds to video frames. Standard LSTM is not an optimal choice for this 4-D input. When fed to fully-connected (FC) or LSTM layers, the 4-D data need to first undertake tensor-to-vector transform, which diminishes all structural information. To avoid losing spatial information, we adopt a recently-proposed network design known as ConvLSTM~\cite{ShiCWYWW15}. It has been successfully applied to the precipitation nowcasting task in Hong Kong. The key idea of ConvLSTM is to implement all operations, including state-to-state and input-to-state transitions, with kernel-based convolutions. This way the 4-D tensors can be directly utilized with spatial structure sustained. In detail, the three gating functions in ConvLSTM are calculated according to the equations below,
\begin{eqnarray}
i_t &=& \sigma\left( \W_{x,i} \otimes \X_{t} + \W_{h,i} \otimes \H_{t-1} \right),  \\
o_t &=& \sigma\left( \W_{x,o} \otimes \X_{t} + \W_{h,o} \otimes \H_{t-1} \right),  \\
f_t &=& \sigma\left( \W_{x,f} \otimes \X_{t} + \W_{h,f} \otimes \H_{t-1} \right),
\end{eqnarray}
where we let $\X_t, ~\H_t$ be the input / hidden state at time $t$ respectively. $\W$'s are the kernels to be optimized. $\otimes$ represents spatio-temporal convolution operator.

Investigating previous $\H$ and current $\X$, the recurrent model synthesizes a new proposal for the cell state, namely
\begin{eqnarray}
\widetilde \C_t &=& \tanh \left( \W_{x,c} \otimes \X_{t} + \W_{h,c} \otimes \H_{t-1} \right).
\end{eqnarray}

The final cell state is obtained by linearly fusing the new proposal $\widetilde \C_t$ and previous state $\C_{t-1}$:
\begin{eqnarray}
\C_t = f_t \odot \C_{t-1} + i_t \odot \widetilde \C_t,
\end{eqnarray}
where $\odot$ denotes the Hadamard product. To continue the recurrent process, it also renders a filtered new $\H$:
\begin{eqnarray}
\H_t = o_t \odot \tanh ( \C_{t-1} ) .
\end{eqnarray}

We would emphasize that most variables are still 4-D tensors, whose sizes can be inferred from context. Because $\H$ is slightly smaller than $\X$ ($14 \times 8 \times 64 \times 10$ v.s. $16 \times 10 \times 64 \times 10$), we pad $\H$ with zeros to equal their sizes before the computations.

\subsection{Steering-Predicting Sub-network}

\begin{figure*}[t]
\centering
   \includegraphics[width=0.9\linewidth]{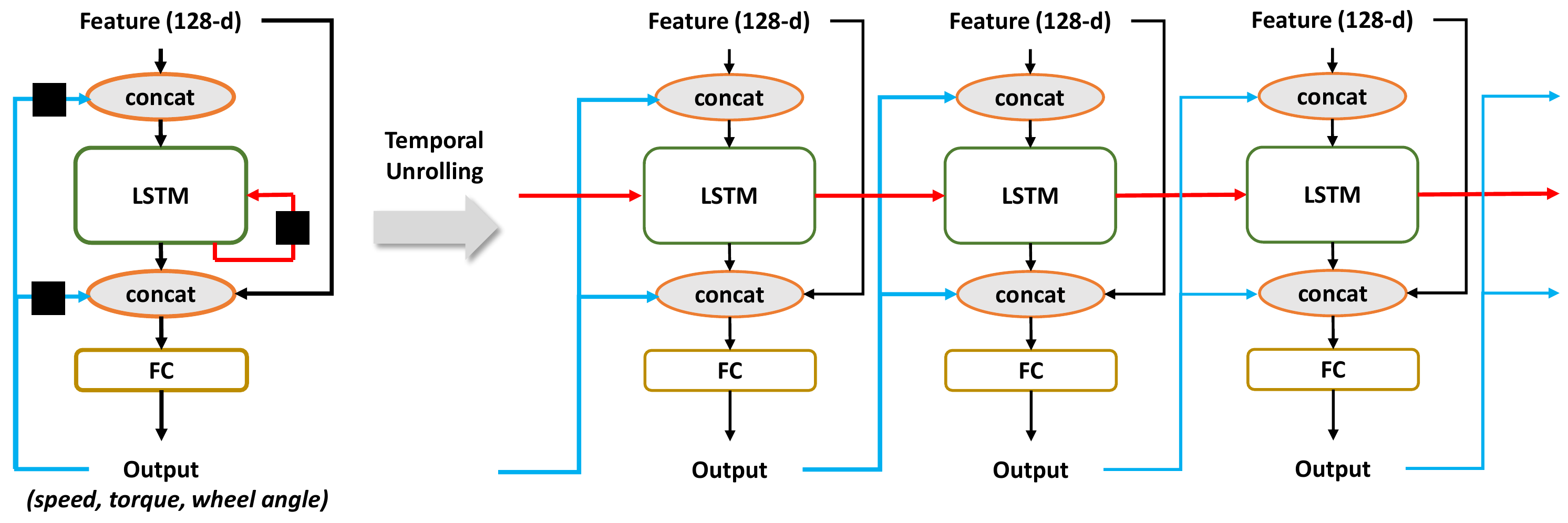}
   \caption{The steering-predicting sub-network. Black blocks on the left diagram indicates recurrence with 1-step delay. To illustrate the temporal dynamics, we use graph unrolling for three consecutive time steps.}
\label{fig:rnn}
\end{figure*}

Fig.~\ref{fig:rnn} depicts our proposed steering-predicting sub-network. It fuses several kinds of temporal information at multiple network layers.

\subsubsection{Temporal Fusion and Recurrence}

As shown in Fig.~\ref{fig:rnn}, There are totally three recurrences during network forwarding computation. One may observe an LSTM unit in the core of this sub-network. It admits a 1-step recurrence, namely forwarding its time-$t$ state to time $t+1$. In fact, the input to this LSTM is not only the 128-d feature vector, which is extracted from the sub-network as described in Sec.~\ref{subsec:fen}. We propose to concatenate previous steering actions and vehicle status with this 128-d vector. To this end, we add another 1-step recurrence between the final output (namely the layer predicting vehicle speed, torque and wheel angle) and the ``concat" layers right before / after LSTM. The concat layer before LSTM append previous vehicle speed, torque and wheel angle to the 128-d extracted feature vector, forming a 131-d vector. The concat layer right after the LSTM layer is comprised of 128-d extracted feature vector + 64-d LSTM output + 3-d previous final output. The major benefit of two concat layers is fully exploiting temporal information.

\begin{table*}[t]
\caption{Key information of the data which benchmarked Udacity self-driving challenge 2. \cmark and \xmark~ indicates the corresponding information is recorded or not.}
\centering
\begin{tabular}{c|c|r|c|c|c|c|c|c}
  \hline\hline
 \textbf{Clip Name} & \textbf{Collection Date} & \textbf{Frame Count}   & \textbf{GPS} & \textbf{Speed} & \textbf{Torque} & \textbf{Wheel} & \textbf{Camera} &\textbf{Use Type}    \\
  \hline
 \textbf{Udacity Dataset 2-3 Compressed} &  2016-10-10 & 223988    &  \cmark & \cmark & \cmark & \cmark & L/M/R & Train \\
  \hline
  \textbf{Challenge 2 \& 3: EI Camino Training Data} & 2016-10-25 & 147120 &  \cmark & \cmark & \cmark & \cmark & L/M/R & Train  \\
 \hline
 \textbf{Ch2\_002: Udacity Self Driving Car} & 2016-11-17 & 33808  &  \cmark & \cmark & \cmark & \cmark & L/M/R & Train \\
 \hline
 \textbf{Ch2\_001: Udacity Self Driving Car} & 2016-11-18 & 5614  &  \xmark & \xmark & \xmark & \cmark & M & Test \\
 \hline
\end{tabular}
\label{table:data}
\end{table*}

\subsubsection{The Overall Multi-Task Objective}


We define a loss term for each of three kinds of predictions, namely $\mathcal{L}_{steer}, \mathcal{L}_{torque}$ and $\mathcal{L}_{speed}$. Each of them is an average of per-frame loss. Recall that $\mathcal{L}_{steer}$ is defined in Eqn.~(\ref{eqn:steer}). Likewise, we define $\mathcal{L}_{torque}$ and $\mathcal{L}_{speed}$.

The final objective function is as following:
\begin{eqnarray}
\mathcal{J} = \gamma \mathcal{L}_{steer} + \mathcal{L}_{speed} + \mathcal{L}_{torque}, \label{eqn:obj}
\end{eqnarray}
where $\gamma$ is introduced to emphasize wheel angle accuracy and in practice we set $\gamma=10$.


\section{Experiments}
\label{sec:experiment}

\subsection{Dataset Description}

\begin{figure*}[t]
\centering
   \includegraphics[width=0.85\linewidth]{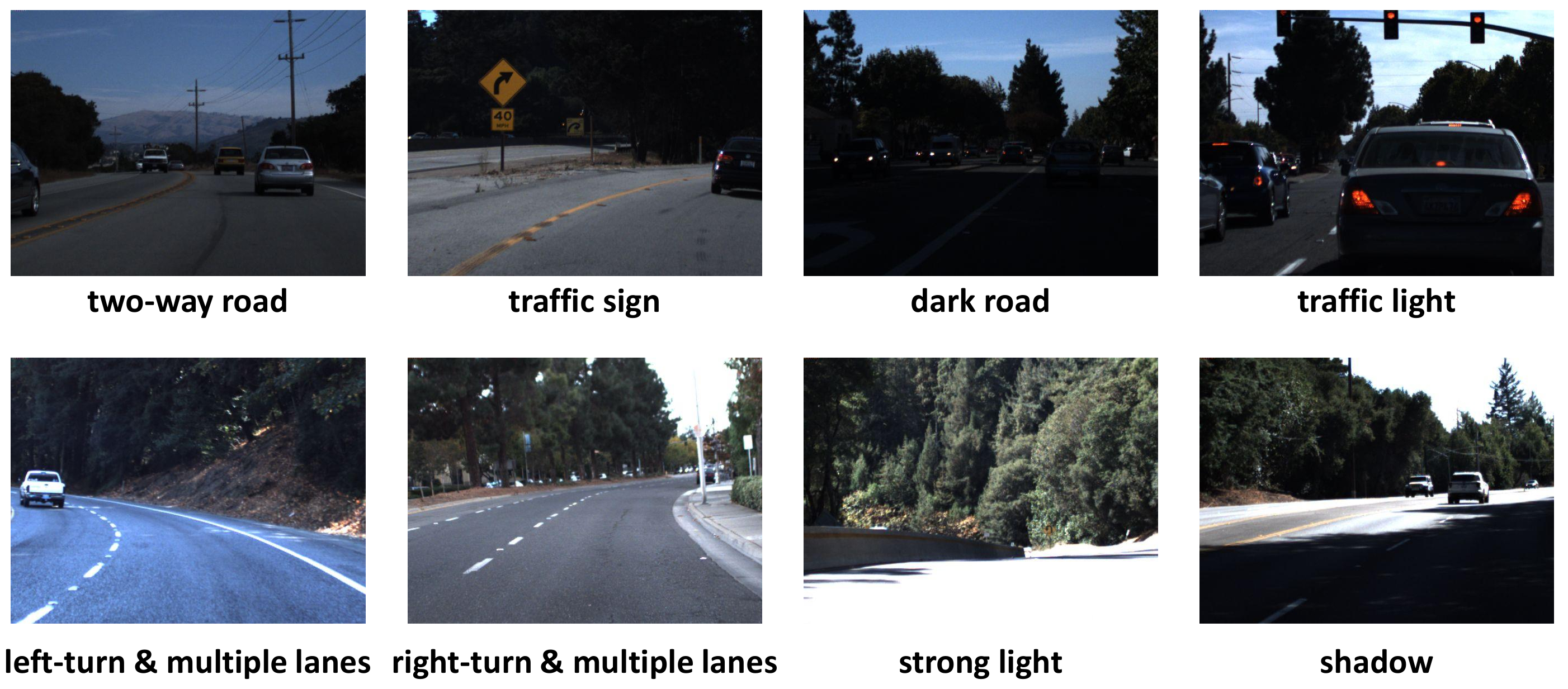}
   \caption{Example video frames in the Udacity dataset.}
\label{fig:example}
\end{figure*}

\begin{figure*}[t]
\centering
   \includegraphics[width=0.95\linewidth]{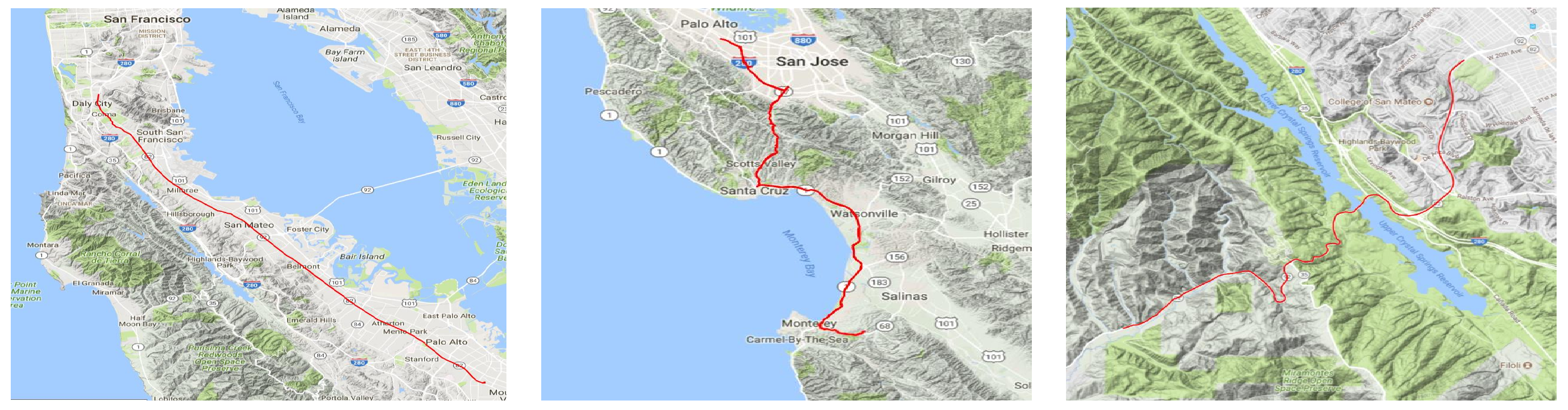}
   \caption{GPS trajectory of three selected video sequences in the Udacity dataset.}
\label{fig:gps}
\end{figure*}

Collecting experimental data for autonomous steering can be roughly cast into three methods, including the logs of human drivers, synthetic data from racing games like Euro Truck or TORCS, or crowd-sourced data uploaded by commercial dash-cameras. After assessing above methods, our evaluations stick to using human logging data. The synthetic data from games come in large volume and with noise-free annotations. However, since the visual surroundings are rendered via computer graphics techniques, the visual appearance often apparently differs from real driving scenes. This may cause severe problem when the developed model is deployed in real cars. On the other hand, crowd-sourced dash-cameras record ego-motion of the vehicles, rather than drivers' steering actions. In other words, what the steering models learn from such data is some indirect indicator rather than the due steering action per se. In addition, the legal risk of personal privacy was not presently well addressed in such data collection.

The company Udacity launched a project of building open-source self-driving cars in 2016 and hosted a series of public challenges\footnote{\url{https://www.udacity.com/self-driving-car}}. Among these challenges, the second one aims to predict real-time wheel steering angles from visual input. The data corpus is still under periodic updating after the challenge. We adopt a bug-free subset of this Udacity dataset for experimental purpose. Table~\ref{table:data} summarizes the key information of the experimental dataset. In specific, videos are captured at a rate of 20 FPS. For each video frame, the data provider managed to record corresponding geo-location (latitude \& longitude), time stamp (in millisecond) and vehicle states (wheel angle, torque, driving speed). Data-collecting cars have three cameras mounted at left / middle / right around the rear mirror. Only the middle-cam video stream is given for the testing sequences, we only use the mid-cam data.

We draw a number of representative video frames from the experimental data. The frames are shown in Fig.~\ref{fig:example}. Moreover, our exposition also includes the GPS trajectories of selected video sequences in Table~\ref{table:data}, which is found in Fig.~\ref{fig:gps}. As seen in above figures, the experimental data is mainly collected on diverse road types and conditions at California, U.S.A. The learned steering model is desired to tackle road traffic, lighting changes, and traffic signs / lights.

\begin{figure*}
\centering
   \includegraphics[width=\linewidth]{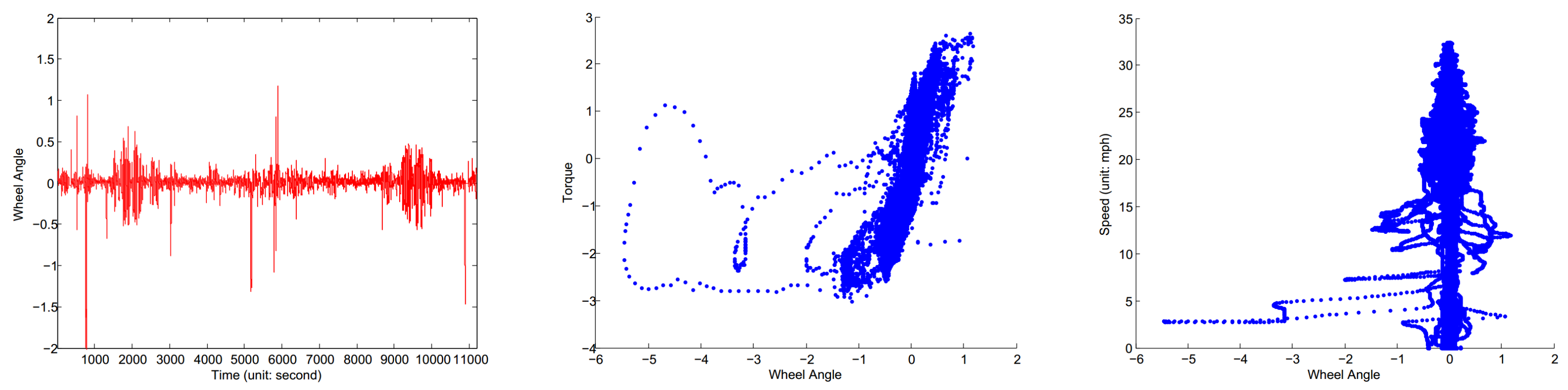}
   \caption{Data statistics of the video clip ``Udacity Dataset 2-3 Compressed". Left: the wheel angels collected from a human driver at different time stamps. Middle: the distribution of (torque, wheel angle) pairs. Right: the distribution of (driving speed, wheel angle) pairs.}
\label{fig:data_stats}
\end{figure*}

Recall that we include three losses in the final prediction (corresponding to driving speed, torque and wheel angle respectively) in the steering-predicting sub-network. This is motivated by the inter-correlation among them. To illustrate it, Fig.~\ref{fig:data_stats} correlates speed v.s. wheel angle, torque v.s. wheel angle, and plots the sequence of wheel angles. Wheel angles tend to zeros on straight roads. The dominating zeros may cause numerical issues when tuning the network parameters. As a step of data taming, we standarize the wheel angles by enforcing a zero-mean and unit standard variation.

\subsection{Network Optimization}


The experiments are conducted on a private cluster with 11 computing nodes and 6 Titan X GPU. All code is written in Google's TensorFlow framework. The following is some crucial parameters for re-implementing our method: dropout with a ratio of 0.25 (in the terminology of TensorFlow, this implies only 25\% neurons are active at specific layer) is used in FC layers. Weight decaying parameter is set to $5 \times 10^{-5}$. Each mini-batch is comprised of 4 15-frame inputs. The learning rate is initialized to $1 \times 10^{-4}$ and halved when the objective is stuck in some plateau. We randomly draw $5\%$ of the training data for validating models and always memorizes the best model on this validation set. For the stochastic gradient solver, we adopt ADAM. Training a model requires about 4-5 days over a single GPU.

To avoid gradient explosion for all recurrent units during training, we clap their stochastic gradients according to a simple rule below:
\begin{eqnarray}
global\_norm = \frac{1}{m} \sqrt{\sum_{i=1}^m \| g_i \|_2^2}
\end{eqnarray}
where $m$ denotes the total number of network parameters and let $g_i$ be the partial gradient of the $i$-th parameter. The real gradient is calculated by
\begin{eqnarray}
g_i = g_i \cdot \frac{clip\_norm}{\max (clip\_norm, global\_norm)},
\end{eqnarray}
where $clip\_norm$ is some pre-set constant. When $global\_norm$ is smaller than $clip\_norm$, no gradient will be affected, otherwise all will be re-scaled to avoid gradient explosion.

%
%

\subsection{Performance Analysis}

\subsubsection{Comparison with Competing Algorithms}

\begin{table}[t]
\caption{Key information of the competing algorithms which benchmarked Udacity self-driving challenge 2. The column ``Memory" and ``Weight" record the estimated memory consumption (in MB) and parameter count of the models respectively.}
\centering
\begin{tabular}{|c|r|r|r|}
  \hline\hline
 \textbf{Model} & \textbf{RMSE} & \textbf{Memory}   & \textbf{Weight}     \\
  \hline
 \textbf{Zero} &  0.2077 & -- & -- \\
 \hline
 \textbf{Mean} &  0.2098 & -- & -- \\
 \hline
 \textbf{AlexNet} &  0.1299 & 5.7339 & 265.7434 \\
 \hline
 \textbf{PilotNet} &  0.1604 & 0.2046 & 0.5919 \\
 \hline
 \textbf{VGG-16} &  0.0948 & 15.3165 & 134.2604 \\
 \hline
 \textbf{ST-Conv + ConvLSTM + \newline LSTM} &  0.0637 & 0.4802 & 37.1076 \\
 \hline
\end{tabular}
\label{table:accuracy}
\end{table}

\begin{figure*}
\centering
   \includegraphics[width=0.95\linewidth]{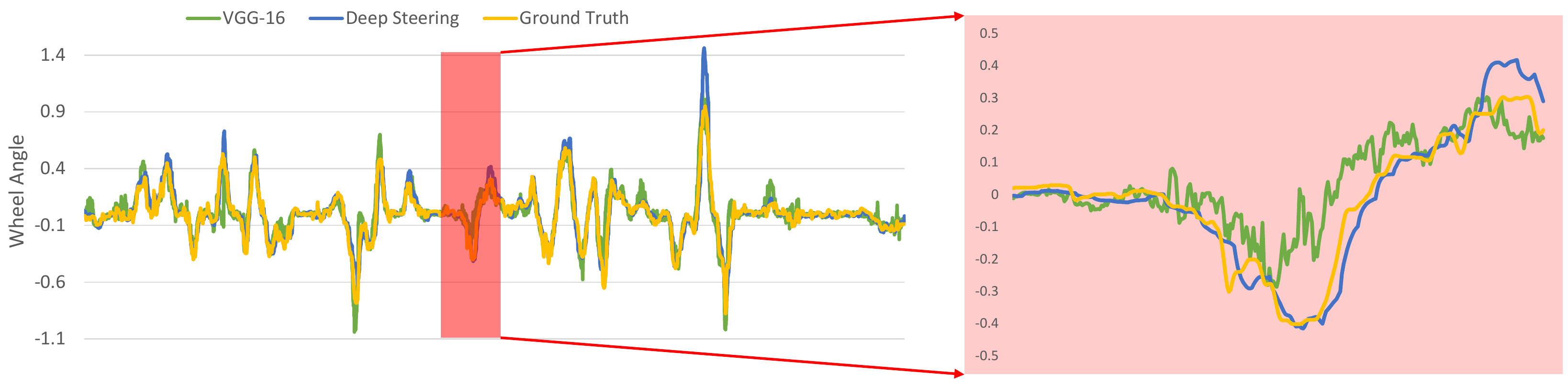}
   \caption{Left: steering wheel angles of the video sequence ``Ch2\_001: Udacity Self Driving Car". We plot the ground truth and the predictions of two models (our proposed model and VGG-16 based model). Right: a selected sub-sequence is highlighted such that more detailed difference can be clearly observed.}
\label{fig:angle_curve}
\end{figure*}

\begin{figure*}
\centering
   \includegraphics[width=0.95\linewidth]{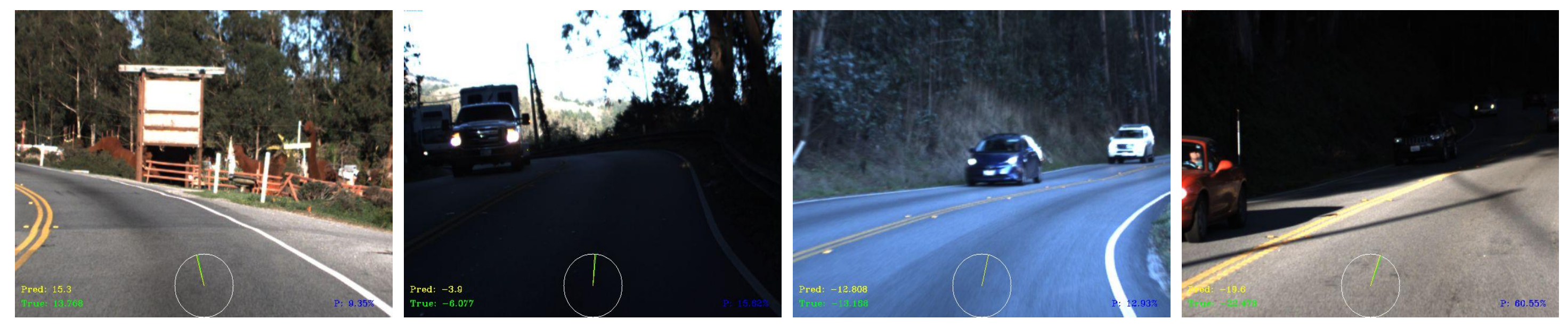}
   \caption{We select for representative scenarios from the testing video sequence. Ground truth steering angles (displayed in green) and our predictions (in yellow) are both imposed on the video frames. Note that our predictions are nearly identical to the ground truth in these challenging inputs.}
\label{fig:angle}
\end{figure*}

We compare the proposed Deep Steering with several competing algorithms. Brief descriptions of these competitors are given as below:
\begin{itemize}
\item \textbf{Zero} and \textbf{Mean}: these two methods represent blind prediction of the wheel angles. The former always predicts a zero wheel angle and the latter outputs the mean angle averaged over all video frames in training set.
\item \textbf{AlexNet}: the network architecture basically follows the seminal AlexNet, with some parameters slightly tailored to the problem that we are considering. We borrow other's AlexNet model pre-trained on the game GTA5 (Grand Theft Auto V), and fine-tune it on the Udacity data.
\item \textbf{PilotNet}: this is the network proposed by NVIDIA. We re-implement it according to NVIDIA's original technical report. All input video frames are resized to $200 \times 88$ (this is the recommended image resolution in NVIDIA's paper) before feeding PilotNet.
\item \textbf{VGG-16}: this network is known to be among the state-of-the-art deep models in the image classification domain. Following prior practical tactics, all convolutional layers are almost freezed during fine-tuning and fully-connected layers are the major target to be adjusted on the Udacity data. Note that both PilotNet and VGG-16 are not recurrent networks and thus ignoring the temporal information.
\end{itemize}

\begin{figure*}
\centering
   \includegraphics[width=0.85\linewidth]{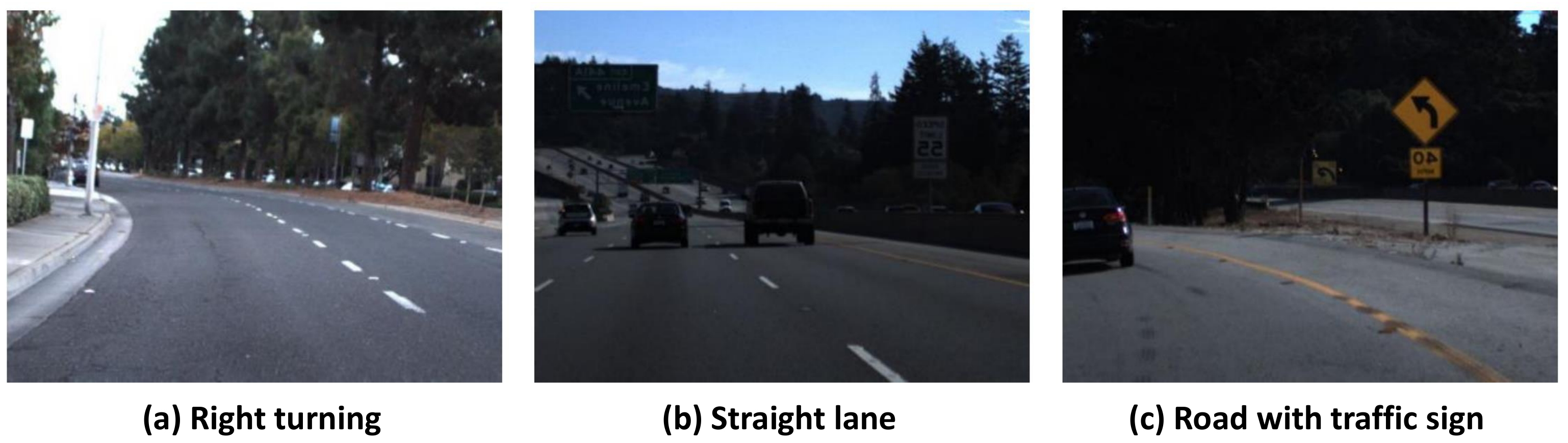}
   \caption{Mirroring the training video frames.}
\label{fig:mirror}
\end{figure*}

Since we mainly focus on predicting the wheel angle, hereafter the model performance will be reported in terms of RMSE (root mean squared error) of wheel angles (namely $\sqrt{\mathcal{L}_{steer}}$) unless otherwise instructed. The model performance are shown in Table~\ref{table:accuracy}, from which we have several immediate observations. First, the design of network heavily correlates to the final performance. Particularly, deeper networks exhibit advantages in representing complex decision function. It is well-known that VGG-16 is a much deeper base network and tends to outperform shallower AlexNet when transferred to other image-related tasks. Our experiments are consistent to this prior belief. Note that the original paper of PilotNet did not report their RMSE value nor the data set used in the road test. Through our re-implementation of PilotNet, we find that PilotNet does not perform as well as other deep models, which may reveal that PilotNet has limitation in deep driving task. Secondly, besides different base models, our model (the last row in Table~\ref{table:accuracy}) clearly differs from others by incorporating temporal information. The experimental results show that ours dominates all other alternatives. We will later show more ablation analysis.

To further investigate the experimental results, Fig.~\ref{fig:angle_curve} plots the wheel angles in a testing video sequence. Specifically, the ground truth collected from human driver and the predictions by VGG-16 and our model are displayed. We also take a sub-sequence which corresponds to some abrupt turnings on the road, for which the wheel angles are plotted on the right sub-figure in Fig.~\ref{fig:angle_curve}. Clearly, VGG-16's predictions (the green curve) are highly non-smooth, which indicates that temporal modeling is key to ensure a smooth driving experience. In Fig.~\ref{fig:angle}, we draw four representative testing video frames and impose the wheel angles of ground truth / our model's prediction.

\subsubsection{Data Augmentation via Mirroring}

\begin{table}[t]
\caption{RMSE loss with mirroring data augmentation.}
\centering
\begin{tabular}{|c|r|r|r|}
  \hline\hline
 \textbf{} & \textbf{Phase-1} & \textbf{Phase-2}   & \textbf{Phase-3}     \\
  \hline
 \textbf{RMSE} &  0.0637 & 0.0698 & 0.0609 \\
 \hline
\end{tabular}
\label{table:mirror}
\end{table}

Deep models are often defined by millions of parameters and have tremendous learning capacity. Practitioners find that data set augmentation is especially helpful, albeit tricky, in elevating the generalization ability of a deep model. A widely-adopted data augmenting scheme is mirroring each image, which doubles the training set. In this work we also explore this idea. Some examples of mirrored video frames are presented in Fig.~\ref{fig:mirror}. We should be aware of the pitfalls caused by the mirroring operation. Though largely expanding the training set, it potentially changes the distribution of real data. For example, Fig.~\ref{fig:mirror}(a) converts a right-turning to left-turning, which violates the right-driving policy and may confuse the driving model. Likewise, the yellow lane marking in Fig.~\ref{fig:mirror}(b) changes to be on the right-hand side of the vehicle after mirroring. Since the yellow lane marking is an important cue in steering, the mirrored frame may adversely  affect the performance. In Fig.~\ref{fig:mirror}(c), the traffic sign of speed limit has a mirrored text which is not human understandable.

To avoid potential performance drop, we adopt a three-phase procedure for tuning the network parameters. In phase 1, the network is trained using the original data set. Phase 2 fine-tunes the network on only the mirrored data. And eventually the model is tuned on the original data again. The experimental results in terms of RMSE are shown in Table~\ref{table:mirror}. As seen, it is clearly validated that using augmented data can improve the learned deep model.

\subsubsection{Keyframe reduction}


\begin{table}[t]
\caption{RMSE loss with reduction.}
\centering
\begin{tabular}{|c|r|}
  \hline\hline
 \textbf{Training Set Reduction Scheme} & \textbf{RMSE} \\
  \hline
 \textbf{No Reduction} &  0.0652 \\
 \hline
 \textbf{Top-Region Cropping}& 0.1066 \\
 \hline
 \textbf{Spatial Sub-sampling}& 0.0697 \\
 \hline
 \textbf{Temporal Sub-sampling with a 1/4 factor}& 0.1344 \\
 \hline
 \textbf{Salient Keyframe Only}& 0.0945 \\
 \hline
\end{tabular}
\label{table:reduction}
\end{table}

\begin{table}[t!]
\caption{ablation analysis results.}
\centering
\begin{tabular}{|c|r|}
  \hline\hline
 \textbf{Model} & \textbf{RMSE}  \\
  \hline
  \textbf{Baseline} &  0.0637 \\
 \hline
 \textbf{Without residual aggregation} &  0.1003 \\
 \hline
 \textbf{Without temporal recurrences} &  0.0729 \\
 \hline
 \textbf{Without ConvLSTM} &  0.0697 \\
 \hline
\end{tabular}
\label{table:ablation}
\end{table}

Video frames in the training set total more than one third million. Since each training epoch requires one pass of data reading, reducing the frame count or file size represents effective means for expediting the training process. To this aim, we have empirically exploited various alternatives, including 1) \emph{No Reduction}: the frames remain the original spatial resolution ($640 \times 480$) as provided by Udacity challenge 2 organizers. Deep network parameters are properly adjusted if they are related to spatial resolution (such as convolutional kernel size), otherwise remain unchanged; 2) \emph{Top-Region Cropping}: for an original $640 \times 480$ video frame, its top $640 \times 200$ image region is observed to respond to mostly sky rather than road conditions. It is thus reasonable to crop this top image region, saving the memory consumption; 3) \emph{Spatial Sub-sampling}: we can uniformly resize all video frames to a much lower spatial resolution. The caveat lies in finding a good tradeoff between key image detail preservation and file size. In this experiment we resize all frames to $320 \times 240$; 4) \emph{Temporal Sub-sampling}: the original videos are 20 FPS. A high FPS is not computationally favored since consecutive frames often have similar visual appearance and encode redundancy. We try the practice of reducing FPS to 5 (namely a 1/4 temporal sub-sampling); 5) \emph{Salient Keyframe Only}: indeed, most human drivers exhibit conservative driving behaviors. Statistically, we find that most of wheel angles approach zeros during driving. It inspires our keeping only ``salient" video frames (frames corresponding to $12^o$ or larger wheel angles) and their neighboring frames. This reduces total frames from 404,916 to 67,714.

The experimental evaluations are shown in Table~\ref{table:reduction}. Compared with not doing any reduction, spatial sub-sampling with a proper resizing factor does not affect much the final performance. All other alternatives prove not good choices. Intuitively, top image region typically corresponds to far-sight view which may also contain useful information for driving. And the failure of other two temporal reduction methods may be caused by changing data distribution along the temporal dimension.

\subsubsection{Ablation Analysis}

Our proposed model include several novel designs, such as residual aggregation and temporal modeling. To quantitatively study the effect of each factor, this section presents three ablative experiments, including 1) removing residual aggregation (while keeping all other layers and using defaulted parameters for training). Specifically, we remove the skip connection around the first convolutional layer. The goal is to verify the effect of low-level features on the final accuracy; 2) removing recurrences. In the steering-predicting sub-network, we remove the auto-regressive connection between two ``concat" layers and the final output. This way the LSTM and FC layers have no information about previous vehicle speed, torque or wheel angles; and 3) removing ConvLSTM, namely we evaluate the performance without the ConvLSTM layer in the feature-extracting sub-network, which is supposed to spatially encode historic information.

The results of these evaluation are shown in Table~\ref{table:ablation}. This table shows that low-level features (such as the edge of roads, are indispensable), previous vehicle state(speed, torque and wheel angle) and spatial recurrence are all providing crucial information for the task that we are considering.

\subsection{Visualization}

\begin{figure}
\centering
   \includegraphics[width=\linewidth]{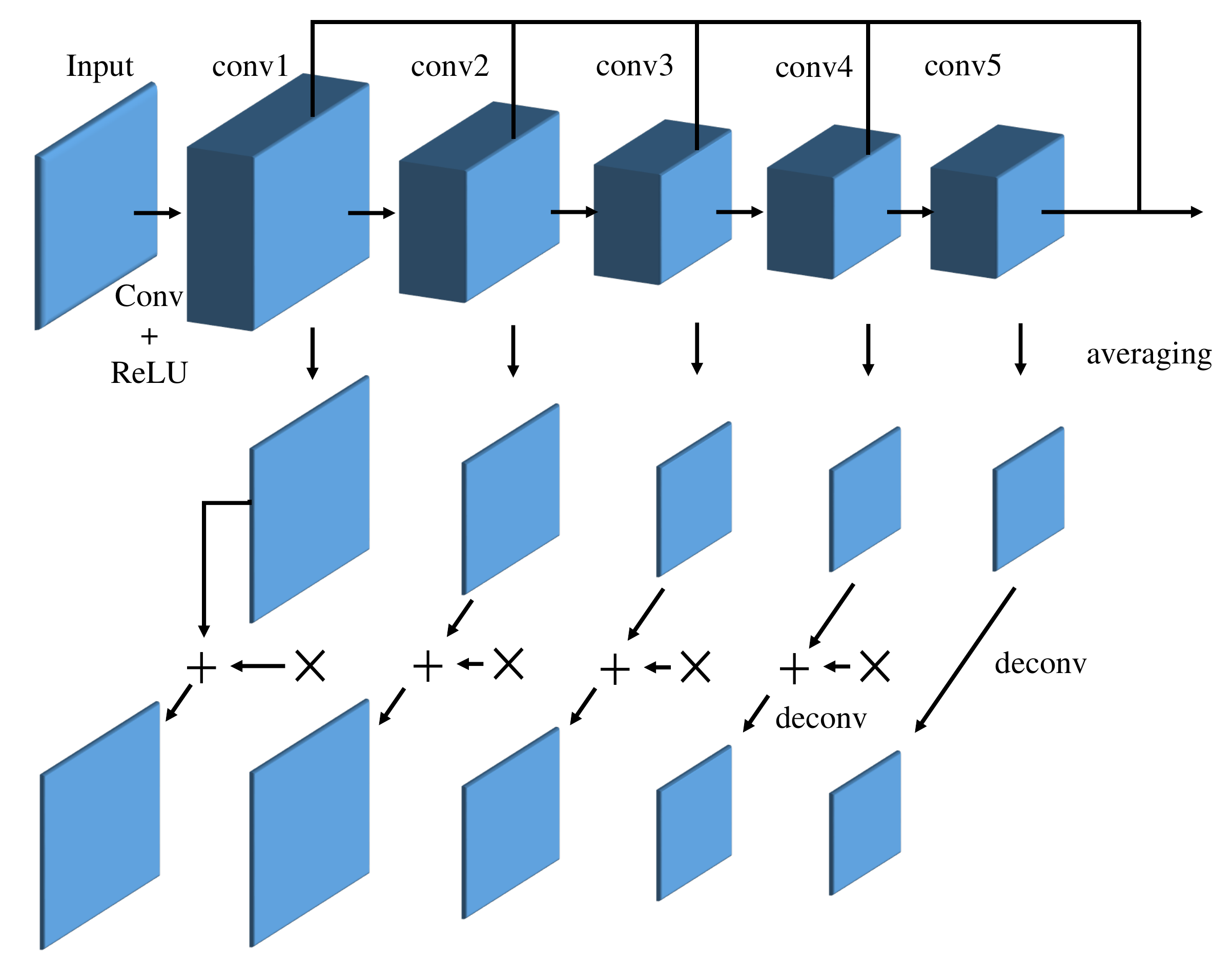}
   \caption{The network used for key factor visualization.}
\label{fig:bvp_structure}
\end{figure}

\begin{figure*}
\centering
   \includegraphics[width=\linewidth]{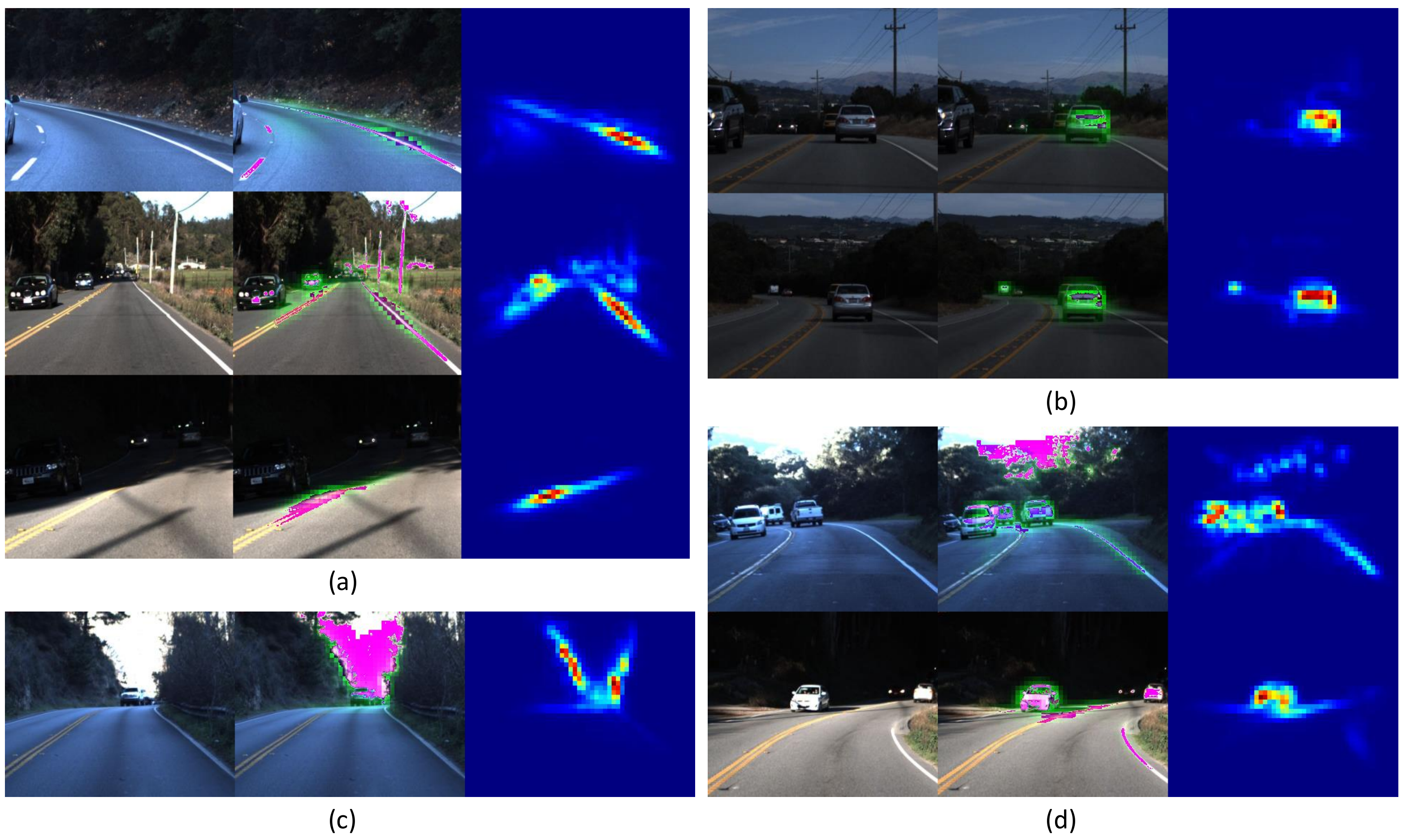}
   \caption{Salient image region detection through visual back-propagation (VBP). The sub-figures (a)(b)(c)(d) represent different types of salient image regions discovered by VBP. For example, (a)(b) find the lane markings and nearby vehicles respectively.  In all sub-figures, the left / middle / right columns correspond to the original video frame, frame with salient region highlighted, the heat map generated by VBP respectively.}
\label{fig:vis_2}
\end{figure*}

\begin{figure*}
\centering
   \includegraphics[width=\linewidth]{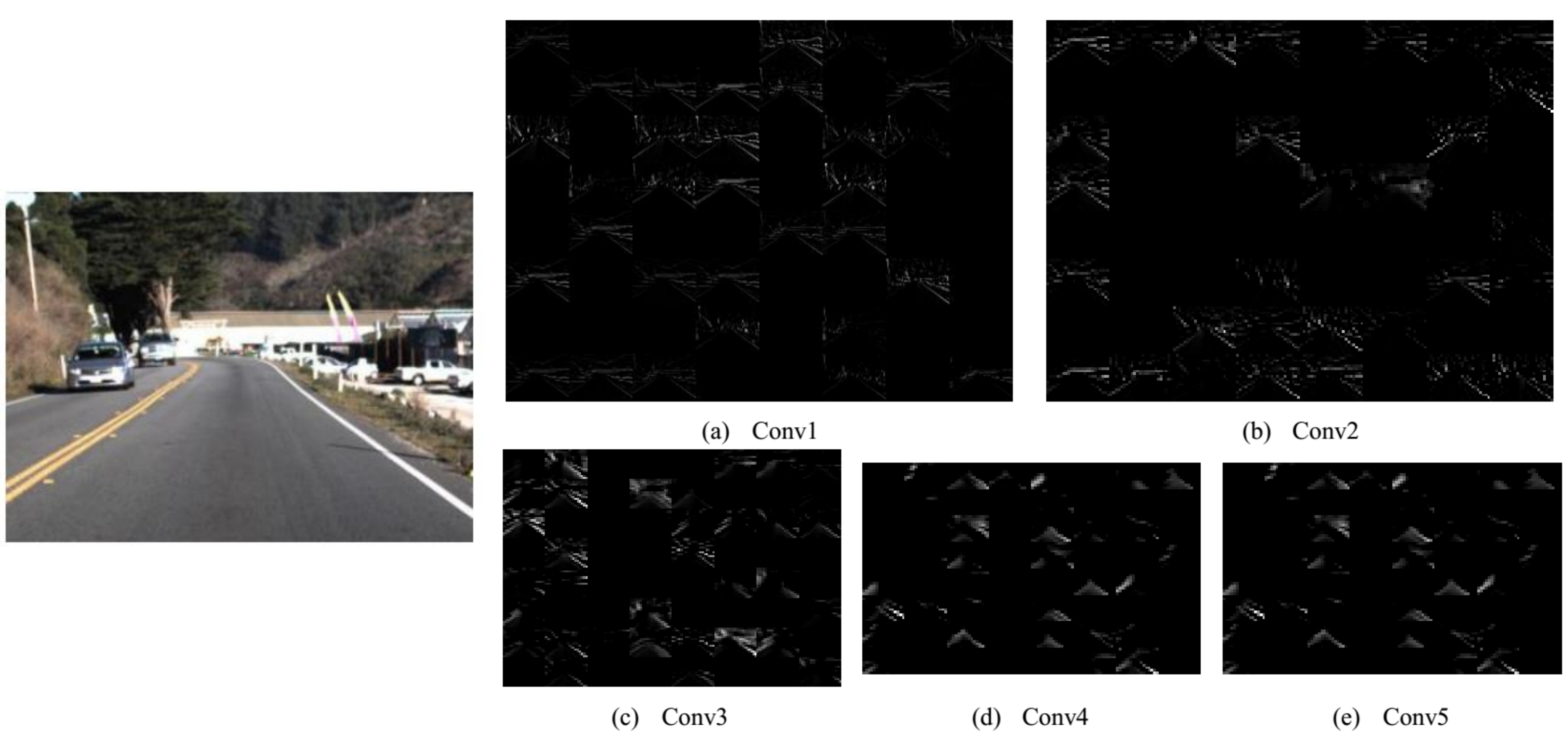}
   \caption{Left: a selected video frame. Right: the response maps at different convolutional layers. ``Conv5" corresponds to the ConvLSTM layer in the feature-extracting sub-network.}
\label{fig:vis_1}
\end{figure*}

Autonomous driving always regards safety as a top priority. Ideally, the learned model's predictive mechanism should be understandable to human users. In the literature of deep learning, substantial efforts~\cite{ZeilerF14,ZhouKLOT16} were devoted to visualize key evidences in the input image that maximally correlate to the network's final output. In this work, we adopt the visual back-propagation (VBP) framework~\cite{BojarskiCCFJMZ16b} proposed by NVIDIA. VBP represents a general idea and can be applied to a large spectrum of deep models. Briefly speaking, the computation of VBP consists of the following steps: 1) for each convolutional layers, average over all channels, obtaining \emph{mean maps}; 2) upsampling each mean map via de-convolution such that its new spatial resolution is the same to the lower layer; 3) perform point-wise multiplication between the upsampled mean map and the mean map at current layer, obtaining \emph{mask map}; 4) add mask map and mean map to obtain \emph{residual map}, which is exactly what we pursue; 5) iterate above steps backwards until the first layer. We illustrate the VBP architecture used in this work in Fig.~\ref{fig:bvp_structure} and show the visualization in Fig.~\ref{fig:vis_2}. It can be seen that key evidences discovered by VBP include lane markings, nearby vehicles, and informative surroundings (such as the silhouette of the valley in Fig.~\ref{fig:vis_2}(c)).

In addition, Fig.~\ref{fig:vis_1} visualizes the response maps for a randomly-selected video frame at all convolutional layers. It is observed that lane markings and nearby cars cause very strong responses, which indicates that the learned model indeed capture the key factor.

\section{Conclusion}
\label{sec:conclusion}

This work addresses a novel problem in computer vision, which aims to autonomously drive a car solely from its camera's visual observation. One of our major technical contributions lies in a deep network which can effectively combine spatial and temporal information. This way exploits the informative historic states of a vehicle. We argue that such a study is rarely found in existing literature. Besides temporal modeling, we have also explored new ideas such as residual aggregation and spatial recurrence. Putting all together leads to a new autonomous driving model which outperforms all other well-known alternatives on the Udacity self-driving benchmark. However, we should be aware that autonomous steering is still in its very early days and there are a number of challenges to be ironed out before the technique is employed on real cars. For example, this work adopts a behavior reflex paradigm. It would be our future research direction to study optimally combining mediated perception and behavior reflex approaches.



\end{document}